\documentclass[]{fairmeta}
\title{CoDrift: Compositional Drifting for Offline Reinforcement Learning}

\author[]{Xiewei Ni}
\author[]{Ruofeng Mei}
\author[*]{Xiangyu Xu}

\affiliation{Xi'an Jiaotong University}

\contribution[*]{Corresponding author}

\abstract{
Offline reinforcement learning is intrinsically multi-objective: a policy must remain compatible with the behavioral support of a fixed dataset while preferentially selecting high-value actions.
We recast these objectives in a common form by viewing each as an action-space motion field that specifies how generated actions should move. This perspective enables heterogeneous learning objectives to be combined directly through field composition.
Inspired by drifting models, we propose \textbf{CoDrift}, a compositional framework for one-step generative policy learning. CoDrift combines three objective-level fields into a unified policy field.
The conditional field preserves state-dependent behavioral structure, while the marginal field pools actions across states to provide a more stable generative signal in the single-positive-sample regime of continuous-control offline RL. The value field moves generated actions toward higher-value regions. 
The composed field is absorbed into a stochastic generator that produces an action with a single forward pass at deployment.
We evaluate CoDrift on 73 tasks from OGBench and D4RL in both offline and offline-to-online settings. CoDrift compares favorably with state-of-the-art methods and achieves the best average rank in both settings.
}

\usepackage{amsmath,amssymb,amsthm}

\definecolor{fqlblue}{HTML}{598BE7}
\definecolor{deflowgreen}{HTML}{2E8B57}
\definecolor{CoDriftred}{HTML}{F54254}
\newcommand{\E}{\mathbb{E}}
\newcommand{\sg}{\operatorname{sg}}

\newcommand{\scorecell}[2]{%
\ensuremath{#1{\scriptscriptstyle\,\pm #2}}%
}

\newcommand{\bestscorecell}[2]{%
\ensuremath{\mathbf{#1}{\scriptscriptstyle\,\pm #2}}%
}

\newcommand{\meanscore}[1]{%
\ensuremath{#1}%
}

\newcommand{\bestmeanscore}[1]{%
\ensuremath{\mathbf{#1}}%
}

\newcommand{\otocell}[4]{%
\ensuremath{%
#1{\scriptscriptstyle\,\pm #2}
\,\rightarrow\,
#3{\scriptscriptstyle\,\pm #4}%
}%
}

\newcommand{\bestotocell}[4]{%
\ensuremath{%
#1{\scriptscriptstyle\,\pm #2}
\,\rightarrow\,
\mathbf{#3}{\scriptscriptstyle\,\pm #4}%
}%
}

\newcommand{\otomean}[2]{%
\ensuremath{#1\,\rightarrow\,#2}%
}

\newcommand{\bestotomean}[2]{%
\ensuremath{#1\,\rightarrow\,\mathbf{#2}}%
}

\newcommand{\fqlhead}{%
\textcolor{fqlblue}{\texttt{FQL}}%
}

\newcommand{\deflowhead}{%
\textcolor{deflowgreen}{\texttt{DeFlow}}%
}

\newcommand{\ourshead}{%
\textcolor{CoDriftred}{\texttt{CoDrift}}%
}

\newcommand{\pendingcell}{\textemdash}

\hypersetup{
  pdftitle={CoDrift: Compositional Drifting for Offline Reinforcement Learning},
  pdfauthor={Xiewei Ni, Ruofeng Mei, Xiangyu Xu},
}

\begin{document}

\maketitle

\section{Introduction}
\begin{figure}[!t]
\centering
\includegraphics[
    width=0.64\textwidth,
    trim=105 45 153 44,
    clip
]{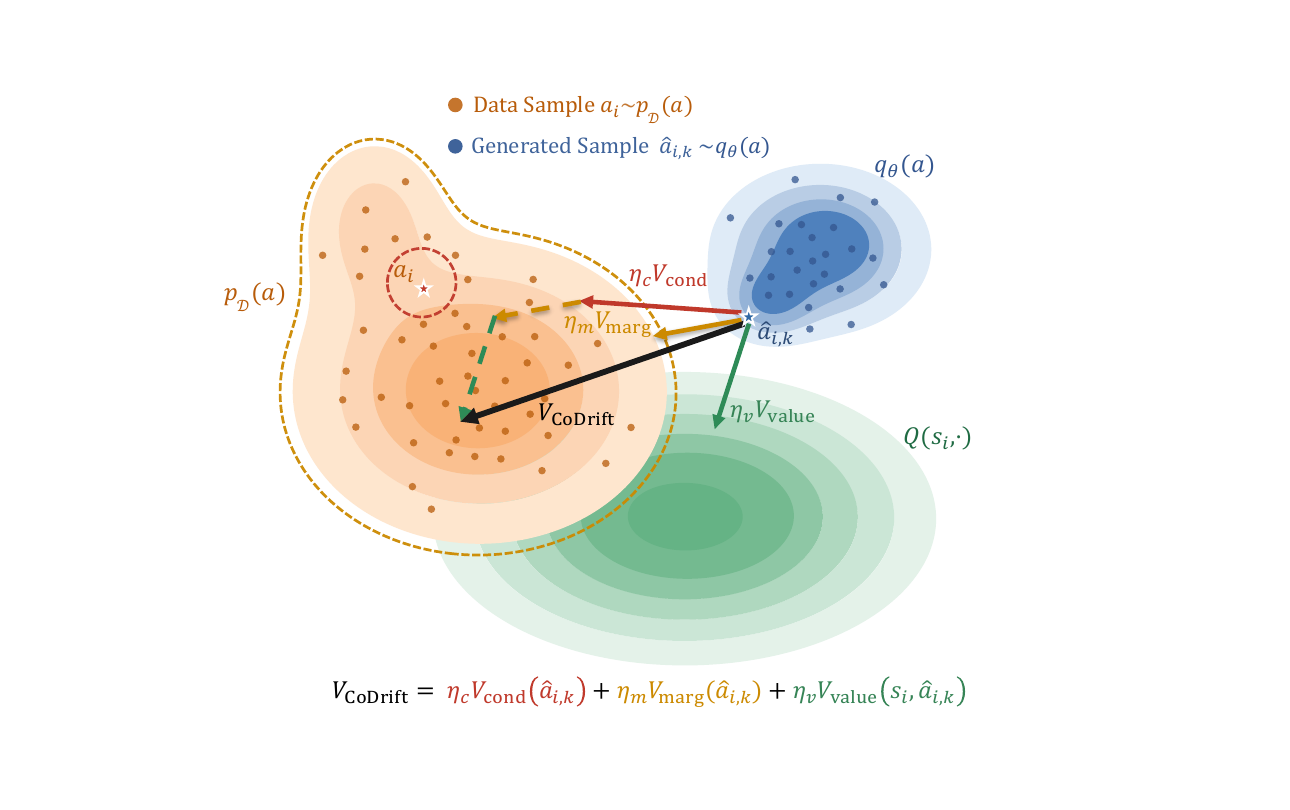}
\caption{
    \textbf{Overview of CoDrift.}
    Conditional and marginal behavioral fields keep generated actions within the support of the offline data at two statistical scales, while the value field moves them toward higher-return regions. The three are composed additively into a single policy field $V_{\mathrm{CoDrift}}$.
}
\label{fig:CoDrift:_field}
\end{figure}

Offline reinforcement learning (RL) aims to learn effective policies entirely from previously collected experience, without further interaction with the environment~\citep{levine2020offline}. 
A central challenge is that policy optimization must satisfy multiple objectives simultaneously. On the one hand, the learned policy should remain compatible with the behavioral support of the offline dataset, since actions far outside the data distribution may lead to unreliable value estimates and poor decisions. On the other hand, merely reproducing the behavior policy is insufficient: the policy must preferentially select high-value actions in order to improve return. Offline RL can therefore be viewed as an intrinsically multi-objective policy-learning problem, balancing behavioral fidelity with value maximization.

Generative policies provide a powerful way to model the behavioral side of this problem. 
In particular, diffusion-based policies have demonstrated good capability to represent complex and multimodal action distributions~\citep{wang2023diffusion,hansenestruch2023idql}, while flow-matching approaches provide an even stronger generative policy class~\citep{park2025flow}. 
These successes suggest that expressive generative modeling is a natural foundation for offline policy learning. 
We build on this perspective, but take a different view of how the multiple objectives of offline RL should be incorporated into a generative policy.

Our starting point is simple: different learning objectives can all be interpreted as specifying how a generated action should move in action space. 
A behavioral objective moves generated actions toward regions supported by the offline data, whereas a value objective moves them toward directions of higher predicted return. 
Once expressed as action-space displacements, these heterogeneous objectives share a common representation and can thereby be composed naturally.
This motivates the central principle of our approach: offline policy learning can be formulated as the \emph{composition of action-space motion fields}, with each field corresponding to a distinct learning objective. 

Our formulation is inspired by drifting models \citep{deng2026drifting}, a recent class of generative models that learns stochastic generators from displacement fields acting directly on generated samples. 
In the original drifting formulation,
a displacement field is constructed from two elementary components: attraction toward positive data samples and repulsion from negative generated samples. 
For our setting, it suggests a broader principle: while the original attraction and repulsion components jointly realize a single distribution-matching objective, different learning objectives can themselves be represented as displacement fields and composed directly in action space. 

Based on this principle, we propose \textbf{CoDrift}, a compositional drifting framework for offline RL. CoDrift instantiates three complementary action-space fields as illustrated in Figure~\ref{fig:CoDrift:_field}. First, a \textbf{conditional behavioral field} preserves state-dependent behavioral support by attracting candidate actions toward actions observed at the corresponding states while maintaining stochasticity through repulsion among generated actions. Second, a \textbf{marginal behavioral field} captures global structure in the action distribution by pooling data and generated actions across states within a minibatch. 
This marginal field is particularly useful in continuous-control offline RL, where essentially every state is paired with only a single observed action. 
Consequently, the conditional field is estimated from a single Monte Carlo sample at each state, whereas marginal field can exploit multiple actions across the batch, yielding a more stable, lower-variance generative signal.
Finally, a \textbf{value field} pushes generated actions toward higher-value regions using action gradients from learned critics. 
These fields play complementary roles and are combined additively into a single generative policy field.

Our contributions are summarized as follows:
\begin{itemize}
    \item We introduce CoDrift, a compositional drifting framework for generative policy learning in offline RL, 
    where heterogeneous learning objectives are represented as additive action-space fields. 
    CoDrift achieves expressive stochastic generation while requiring only a single forward pass at deployment.
    \item We identify the single-positive-sample problem in conditional drifting for offline RL. 
    We address this issue with marginal drifting, which pools actions across states to provide a more stable training signal.
    \item We conduct extensive experiments across 73 tasks from OGBench and D4RL in both offline and offline-to-online settings. 
    CoDrift compares favorably with state-of-the-art approaches, achieving the best average rank in both settings.
\end{itemize}

\section{Preliminaries}

\subsection{Offline RL} We consider a Markov decision process (MDP) $\mathcal{M}=(\mathcal{S},\mathcal{A},P,r,\rho,\gamma)$ \citep{levine2020offline}, where $\mathcal{S}$ is the state space, and $\mathcal{A}=[-1,1]^d$ is the $d$-dimensional continuous action space. $P(\cdot\mid s,a):\mathcal{S}\times\mathcal{A}\to\Delta(\mathcal{S})$ is the transition dynamics, where $\Delta(\mathcal{X})$ denotes the set of probability distributions over a space $\mathcal{X}$.
The reward function is $r:\mathcal{S}\times\mathcal{A}\to\mathbb{R}$, $\rho\in\Delta(\mathcal{S})$ is the initial state distribution, and $\gamma\in[0,1)$ is the discount factor. A policy $\pi_\theta(\cdot\mid s):\mathcal{S}\to\Delta(\mathcal{A})$ induces a trajectory distribution $p^{\pi_\theta}(\zeta)$ under the dynamics $P$ and the initial state distribution $\rho$, and is evaluated by its expected discounted return
\begin{equation}\label{eq:return}
J(\pi_\theta)
=\E_{\zeta\sim p^{\pi_\theta}}
\left[\sum_{t=0}^{\infty}\gamma^t r(s_t,a_t)\right],
\end{equation}
with associated action-value function
\begin{equation*}
Q^{\pi_\theta}(s,a)
=\E_{\zeta\sim p^{\pi_\theta}}
\left[\left.\sum_{t=0}^{\infty}\gamma^t r(s_t,a_t)\,\right|\,s_0=s,a_0=a\right]. 
\end{equation*}

Offline RL seeks to maximize Eq.~\ref{eq:return} using only a fixed dataset $\mathcal{D}=\{(s,a,r,s')\}$ of transitions collected by an unknown behavior policy $\pi_\beta$, without further interaction with the environment. 
    In this work, we also consider the offline-to-online setting, where the offline pre-trained policy is further fine-tuned with a modest amount of online environment interactions.
A fundamental difficulty in offline RL is distribution shift: because $\mathcal{D}$ covers only state--action regions visited by $\pi_\beta$, a critic fit on $\mathcal{D}$ cannot be expected to extrapolate reliably outside the support of the behavior distribution,
and unconstrained value maximization against such a critic drives the actor toward out-of-distribution actions \citep{levine2020offline}.

\paragraph{Behavior-regularized actor--critic methods.}  \citet{wu2019behavior,fujimoto2021minimalist,tarasov2023revisiting} address this issue by balancing value maximization with an explicit constraint that keeps the learned policy close to the offline behavior.
A critic $Q_\varphi$ is typically trained by Bellman regression,
\begin{equation*}
\mathcal{L}_{Q}(\varphi)
=\E_{\substack{(s,a,r,s')\sim\mathcal{D},\\a'\sim\pi_\theta(\cdot\mid s')}}
\left[\left(Q_\varphi(s,a)-r-\gamma Q_{\bar\varphi}(s',a')\right)^2\right], 
\end{equation*}
where $Q_{\bar\varphi}$ denotes a target network \citep{mnih2015human}. 
The actor is then commonly optimized using an objective of the form
\begin{equation*}
\mathcal{L}_{\pi}(\theta)
=\E_{\substack{s\sim\mathcal{D},\\a\sim\pi_\theta(\cdot\mid s)}}
\left[-Q_\varphi(s,a)+\alpha\mathcal{R}(\pi_\theta)\right],
\end{equation*}
where $\mathcal{R}(\pi_\theta)$ regularizes the policy toward the offline behavior distribution and $\alpha$ controls the trade-off between behavioral fidelity and value maximization.

CoDrift follows the same fundamental principle of balancing behavioral fidelity and value maximization, but realizes it in a different form. 
Rather than combining a critic objective with an explicit behavioral penalty at the loss level, CoDrift represents the learning objectives as action-space displacement fields and composes them into a joint policy field, as introduced in the following sections.

\subsection{Drifting Models} \label{sec:drifting_model}
Drifting models \citep{deng2026drifting} are a recent approach to one-step generative modeling. Let $p:=p_{\mathrm{data}}$ be the target distribution on $\mathbb{R}^d$ and $p_\varepsilon:=\mathcal{N}(0,I)$ a prior on $\mathbb{R}^m$. A one-step generator $f_\theta:\mathbb{R}^m\to\mathbb{R}^d$ maps noise $\varepsilon\sim p_\varepsilon$ to a sample $x=f_\theta(\varepsilon)$, inducing the pushforward distribution $q_\theta:=[f_\theta]_{\#}p_\varepsilon$, and the goal is to make $q_\theta$ match $p$. 
Unlike diffusion and flow models, which transport samples through a sequence of intermediate states at inference time, drifting models shift this iterative transport process to training. The generator itself remains a single forward mapping, while its induced distribution $q_\theta$ is progressively moved toward $p$ over the course of optimization.

The transport direction is given by a distribution-dependent \emph{drifting field} $V_{p,q}(x)=V_p^+(x)-V_q^-(x)$, whose two components are kernel-normalized mean-shift vectors:
\begin{equation}\label{eq:drift-field}
\begin{aligned}
V_p^+(x)&=\frac{\E_{u^+\sim p}[\kappa_\tau(x,u^+)(u^+-x)]}{\E_{u^+\sim p}[\kappa_\tau(x,u^+)]},\\
V_q^-(x)&=\frac{\E_{u^-\sim q}[\kappa_\tau(x,u^-)(u^--x)]}{\E_{u^-\sim q}[\kappa_\tau(x,u^-)]},
\end{aligned}
\end{equation}
where $\kappa_\tau(\cdot,\cdot)$ is a positive similarity kernel. Throughout this work, we adopt the Laplace kernel $\kappa_\tau(x,u)=\exp(-\|x-u\|_2/\tau)$, which assigns larger weights to nearby samples, with $\tau$ controlling the effective neighborhood size. 
Because the kernel weights are normalized, each term in Eq.~\ref{eq:drift-field} represents a displacement vector from $x$ to the weighted centroid of a reference set. 
Accordingly, $V_p^+$ acts as an \emph{attractive component}, pulling generated samples toward the target distribution, whereas subtracting $V_q^-$ produces a \emph{repulsive component} that pushes generated samples away from one another and discourages mode collapse.

An important property of the drifting field is its antisymmetry: $V_{p,q}(x)=-V_{q,p}(x)$.  
In particular, when $q=p$, the attractive and repulsive components cancel and $V_{p,q}(x)=0$, so matching the target distribution corresponds to an equilibrium of the drifting dynamics.
Training moves the generator toward this equilibrium by constructing a displaced target for each generated sample. Given $x=f_\theta(\varepsilon)$, the target is $\tilde{x}=x+V_{p,q}(x)$, which is treated as fixed when updating the generator. 
The resulting regression objective is
\begin{equation}\label{eq:drift-loss}
\mathcal{L}_{\mathrm{drift}}(\theta;p,q)
=\E_{\varepsilon\sim p_\varepsilon}\Bigl[
\bigl\|f_\theta(\varepsilon)-\sg\left(f_\theta(\varepsilon)+V_{p,q}(f_\theta(\varepsilon))\right)\bigr\|_2^2\Bigr],
\end{equation}
where $\sg(\cdot)$ is the stop-gradient operator. 
Thus, rather than explicitly integrating the drifting field at inference time, training repeatedly regresses the one-step generator toward samples displaced by the current field.

In practice, the expectations in Eq.~\ref{eq:drift-field} are approximated with minibatch samples. 
Given references $u_1,\ldots,u_N$, each component takes the form $\sum_{i=1}^{N} w_i(u_i-x)$ with normalized weights $w_i=\kappa_\tau(x,u_i)/\sum_{j=1}^{N}\kappa_\tau(x,u_j)$. 
For the repulsive component, the references are other samples generated in the same training step, with the query sample excluded from its own reference set.

For later use, we write $V(x;\mathcal{Y},\mathcal{Z})$ for the minibatch drifting field evaluated at $x$, with positive reference set $\mathcal{Y}$ and negative reference set $\mathcal{Z}$, which can be seen as a Monte Carlo approximation of $V_{p,q}(x)$. Throughout, a query particle is excluded from its own negative references, allowing $\mathcal{Z}$ to denote the full set of generated particles without additional notation.

\section{Method}
We introduce \textbf{CoDrift}, a compositional drifting framework for offline reinforcement learning. 
CoDrift represents different learning objectives as displacement fields acting directly on generated actions and composes them into a single policy field. Specifically, CoDrift contains three objective-level fields: a \emph{conditional behavioral field}, a \emph{marginal behavioral field}, and a \emph{value field}. 
The first two preserve behavioral structure at complementary statistical scales, while the third pushes generated actions toward higher-value regions. Figure~\ref{fig:CoDrift:_field} provides an overview.

We parameterize the policy as a one-step stochastic generator $f_\theta$. 
Given a state $s$ and noise $\varepsilon\sim p_\varepsilon=\mathcal{N}(0,I)$, 
\begin{equation} \label{eq:policy} 
\hat a = f_\theta(s,\varepsilon), \quad \pi_\theta(\cdot\mid s) = \bigl(f_\theta(s,\cdot)\bigr)_{\#}p_\varepsilon , 
\end{equation} 
where $f_\theta$ uses $\tanh$ to ensure that $\hat a\in[-1,1]^d$. Sampling from $\pi_\theta$ therefore requires only one noise draw and one forward pass.

\subsection{Conditional Behavioral Field} 

The conditional behavioral field preserves the state-dependent action distribution of the offline data. At the population level, it instantiates the drifting field of Sec.~\ref{sec:drifting_model} between the behavioral conditional distribution $p_{\mathcal D}(a\mid s)$ and the policy $\pi_\theta(a\mid s)$, where $p_{\mathcal D}(a\mid s)$ plays the role of the target distribution $p_\mathrm{data}$, while $\pi_\theta(a\mid s)$ plays the role of the generated distribution $q_\theta$.

Given a minibatch $\mathcal B=\{(s_i,a_i)\}_{i=1}^{B}$, we draw $K$ independent noise samples for each state, \begin{equation*} 
\varepsilon_{i,k}\sim\mathcal N(0,I), \qquad \hat a_{i,k}=f_\theta(s_i,\varepsilon_{i,k}), \qquad k=1,\ldots,K. 
\end{equation*} 
For each state $s_i$, the positive and generated reference sets are 
\begin{equation} \label{eq:cond_references}
\mathcal Y_i^{\mathrm{cond}}=\{a_i\}, \qquad \mathcal Z_i^{\mathrm{cond}} = \{\hat a_{i,1},\ldots,\hat a_{i,K}\}.
\end{equation} 
Using the minibatch drifting notation introduced in Sec.~\ref{sec:drifting_model}, we define \begin{equation*} 
V_{\mathrm{cond}}(\hat a_{i,k}) := V\!\left( \hat a_{i,k}; \mathcal Y_i^{\mathrm{cond}}, \mathcal Z_i^{\mathrm{cond}} \right). 
\end{equation*}

Because the positive set $\mathcal Y_i^{\mathrm{cond}}$ contains only the single action paired with $s_i$, the attractive component has weight one and reduces to 
\begin{equation} \label{eq:cond-attraction} V_{\mathrm{cond}}^{+}(\hat a_{i,k}) = a_i-\hat a_{i,k}. \end{equation} 
The repulsive component is computed from the other generated actions at the same state, 
\begin{equation*} V_{\mathrm{cond}}^{-}(\hat a_{i,k}) = \sum_{l\neq k} w_{kl}^{\mathrm{cond}} (\hat a_{i,l}-\hat a_{i,k}), 
\end{equation*} 
where the normalized affinities $w_{kl}^{\mathrm{cond}} = \frac{\kappa_\tau(\hat{a}_{i,k},\hat{a}_{i,l})}
{\sum_{j\ne k}\kappa_\tau(\hat{a}_{i,k},\hat{a}_{i,j})}$ following Eq.~\ref{eq:drift-field}, and $\hat a_{i,k}$ is excluded from its own negative references. 
Thus, 
\begin{equation*} V_{\mathrm{cond}} = V_{\mathrm{cond}}^{+} - V_{\mathrm{cond}}^{-} . 
\end{equation*}

A distinctive issue arises in continuous-control offline RL: an exact state is essentially observed only once, and hence $p_{\mathcal D}(a\mid s_i)$ is represented by only one observed action. Consequently, the attractive component in Eq.~\ref{eq:cond-attraction} is effectively a one-sample Monte Carlo estimate of the population conditional drifting field, 
which can have high variance. This motivates a complementary field that can exploit distributional information pooled across states.

\subsection{Marginal Behavioral Field} \label{sec:marginal_field}
The same stochastic generator also induces a marginal action distribution when states are drawn from the offline state distribution.
Recall from Eq.~\ref{eq:policy} that, for a fixed state $s$,
$f_\theta(s,\cdot)$ pushes the noise distribution $p_\varepsilon$ forward to the conditional policy distribution $\pi_\theta(\cdot\mid s)$.
If we additionally draw $s\sim p_{\mathcal D}(s)$, then the joint mapping
\[
(s,\varepsilon)\mapsto f_\theta(s,\varepsilon)
\]
pushes the product distribution
$p_{\mathcal D}(s)p_\varepsilon(\varepsilon)$
forward to the policy-induced marginal action distribution
\begin{equation*}
q_\theta
=
(f_\theta)_{\#}
\bigl(
p_{\mathcal D}\otimes p_\varepsilon
\bigr), 
\end{equation*}
where $\otimes$ denotes product distribution.
Equivalently,
\begin{equation*}
q_\theta(a)
=
\int
p_{\mathcal D}(s)\,
\pi_\theta(a\mid s)\,ds.
\end{equation*}
The corresponding marginal action distribution in the offline data is
\begin{equation*}
p_{\mathcal D}(a)
=
\int
p_{\mathcal D}(s)\,
p_{\mathcal D}(a\mid s)\,ds.
\end{equation*}

This yields a second distribution-matching objective at the marginal level.
In particular, if
\begin{equation}\label{eq:conditional_distribution}
\pi_\theta(a\mid s)
=
p_{\mathcal D}(a\mid s),
\end{equation}
then marginalizing over $s$ immediately gives
\begin{equation} \label{eq:marginal_distribution}
q_\theta(a)
=
p_{\mathcal D}(a).
\end{equation}
Hence matching the marginal action distribution in Eq.~\ref{eq:marginal_distribution} is a necessary condition for matching the conditional behavioral distribution in Eq.~\ref{eq:conditional_distribution}.

More importantly, the marginal distribution is much better sampled in an offline minibatch. For each predicted action $\hat a_{i,k}$, we pool actions across states: 
\begin{equation*} 
\mathcal Y^{\mathrm{marg}} = \{a_1,\ldots,a_B\}, \qquad \mathcal Z_k^{\mathrm{marg}} = \{\hat a_{1,k},\ldots,\hat a_{B,k}\}. 
\end{equation*} 
The marginal behavioral field acting on $\hat a_{i,k}$ is 
\begin{equation*}  V_{\mathrm{marg}}(\hat a_{i,k}) := V\!\left( \hat a_{i,k}; \mathcal Y^{\mathrm{marg}}, \mathcal Z_k^{\mathrm{marg}} \right). 
\end{equation*} 
It decomposes into attractive and repulsive components,
\begin{equation*} V_{\mathrm{marg}} = V_{\mathrm{marg}}^{+} - V_{\mathrm{marg}}^{-}, 
\end{equation*} 
where the attraction is computed from all $B$ offline actions in the minibatch and the repulsion from generated actions pooled across states. 

The crucial difference from conditional drifting is that state--action pairing is deliberately removed at the marginal level.
States used to generate policy actions are sampled from $p_{\mathcal D}(s)$, while positive reference actions are sampled independently from $p_{\mathcal D}(a)$.
Therefore, instead of estimating a conditional expectation from the single action observed at one state ($\mathcal Y_i^{\mathrm{cond}}$ in Eq.~\ref{eq:cond_references}), marginal drifting is able to estimate the corresponding distributional field using $B$ positive samples in $\mathcal Y^{\mathrm{marg}}$. 
Consequently, it provides a more stable, lower-variance generative signal while constraining a distribution that must also match whenever the full conditional behavior is matched. The conditional and marginal fields are thus complementary: the former preserves state--action correspondence, whereas the latter supplies population-level action-distribution information.

\subsection{Value Field} Behavioral matching alone does not solve offline RL: among actions supported by the dataset, the policy should favor those with higher expected return. We therefore express value improvement as a third action-space field. We maintain two critics, $Q_{\varphi_1}$ and $Q_{\varphi_2}$, together with target networks $Q_{\bar\varphi_1}$ and $Q_{\bar\varphi_2}$ obtained through Polyak averaging~\citep{fujimoto2018addressing}. For a transition $(s,a,r,s')\sim\mathcal D$, we sample \begin{equation*} a' = f_\theta(s',\varepsilon'), \qquad \varepsilon'\sim\mathcal N(0,I), \end{equation*} and form the Bellman target \begin{equation*} y = r + \gamma \frac{ Q_{\bar\varphi_1}(s',a') + Q_{\bar\varphi_2}(s',a') }{2}. \end{equation*} The critics minimize \begin{equation*} \mathcal L_{\mathrm{critic}} = \E \left[ \frac{1}{2} \sum_{n=1}^{2} \left( Q_{\varphi_n}(s,a)-y \right)^2 \right]. \end{equation*} When updating the actor, the critics are held fixed. 
Let \begin{equation*} 
Q(s,a) = \frac{ Q_{\varphi_1}(s,a) + Q_{\varphi_2}(s,a) }{2}.
\end{equation*} 
The value field is simply the action gradient of the critic, \begin{equation*} V_{\mathrm{value}}(s,a) = \nabla_a Q(s,a). \end{equation*} It specifies the local action-space direction of increasing predicted return. Unlike the conditional and marginal fields, which arise from distribution matching, the value field is not associated with a reference distribution. Nevertheless, all three objects are displacement fields in the same action space and can therefore be composed directly.

\subsection{Compositional Drifting} \label{sec:joint-objective} 

We now combine the three objective-level fields into a single policy field. For every generated particle $\hat a_{i,k}$, define 
\begin{equation} \label{eq:joint-field}
V_{\mathrm{CoDrift}}(s_i,\hat a_{i,k}) = \eta_c V_{\mathrm{cond}}(\hat a_{i,k}) + \eta_m V_{\mathrm{marg}}(\hat a_{i,k}) + \eta_v V_{\mathrm{value}}(s_i,\hat a_{i,k}),
\end{equation}
where $\eta_c,\eta_m,\eta_v\geq0$ control the relative contributions of conditional behavioral matching, marginal behavioral matching, and value maximization. 
At a finer level, Eq.~\ref{eq:joint-field} contains five displacement components: 
\begin{equation*}
V_{\mathrm{CoDrift}} = \eta_c \left( V_{\mathrm{cond}}^{+} - V_{\mathrm{cond}}^{-} \right) + \eta_m \left( V_{\mathrm{marg}}^{+} - V_{\mathrm{marg}}^{-} \right) + \eta_v V_{\mathrm{value}} .
\end{equation*}
The first four components define two complementary behavioral fields, while the last introduces task-directed value optimization. Similar to the training of drifting models in Eq.~\ref{eq:drift-loss}, the joint field directly defines a displaced target for every generated action: 
\begin{equation*} 
\tilde a_{i,k} = \sg\!\left[ \hat a_{i,k} + V_{\mathrm{CoDrift}}(s_i,\hat a_{i,k}) \right]. 
\end{equation*} 
The actor is finally trained with 
\begin{equation*} 
 \mathcal L_{\mathrm{CoDrift}}(\theta) = \E \left[ \frac{1}{BK} \sum_{i=1}^{B} \sum_{k=1}^{K} \left\| \hat a_{i,k} - \tilde a_{i,k} \right\|_2^2 \right], 
\end{equation*} 
which learns the one-step stochastic policy by moving toward the composed target.

In our implementation, we parameterize the field weights as \begin{equation*}  \eta_v=1, \qquad \eta_c=\alpha\lambda, \qquad \eta_m=\alpha(1-\lambda), \end{equation*} where $\alpha$ controls the overall strength of behavioral regularization and $\lambda\in[0,1]$ balances conditional and marginal behavioral matching. 

\section{Experiments}
We evaluate \textbf{CoDrift} in both offline and offline-to-online reinforcement learning settings across a broad collection of continuous-control benchmarks. Our experiments are designed to answer two main questions: 1) whether compositional drifting provides a competitive generative policy-learning framework across diverse offline RL tasks, and 2) whether marginal drifting improves over conditional drifting alone. 

\subsection{Experimental Setup}

\paragraph{Benchmarks.} We follow the benchmark suite and evaluation setting of existing works~\citep{park2025flow,mu2026deflow}. 
Specifically, we evaluate on OGBench~\citep{park2025ogbench}, which contains diverse long-horizon continuous-control tasks spanning navigation, locomotion, and manipulation. 
The datasets of OGBench are collected by task-agnostic policies, and we use their standard single-task variants, where a fixed evaluation goal is specified, and semi-sparse task rewards are relabeled from the original trajectories.
We consider ten state-based datasets across six domains: AntMaze, HumanoidMaze, AntSoccer, Cube, Scene, and Puzzle, with five evaluation tasks per dataset, together with five pixel-based visual manipulation tasks. 
We additionally evaluate on D4RL~\citep{fu2020d4rl}, including six AntMaze navigation tasks and twelve Adroit manipulation tasks. Altogether, the offline evaluation comprises 73 tasks spanning different state and action dimensions, observation modalities, reward structures, and control problems. For offline-to-online RL, we follow the same 15-task protocol used by FQL and DeFlow~\citep{park2025flow,mu2026deflow}, including five OGBench tasks, six D4RL AntMaze tasks, and four D4RL Adroit tasks. The policy is first trained offline and subsequently fine-tuned using online interactions without changing the learning objective.

\paragraph{Baselines.} We compare CoDrift against eleven representative offline RL methods spanning several policy classes. Gaussian-policy baselines include behavior cloning (BC), Implicit Q-Learning (IQL; \citealp{kostrikov2022offline}), and ReBRAC~\citep{tarasov2023revisiting}. Diffusion-based baselines include IDQL~\citep{hansenestruch2023idql}, SRPO~\citep{chen2024srpo}, and Consistency Actor-Critic (CAC; \citealp{ding2024consistency}). 
Flow-based baselines include FQL~\citep{park2025flow}, DeFlow~\citep{mu2026deflow}, and three flow-policy variants introduced as baselines by \citet{park2025flow}: FAWAC, FBRAC, and IFQL, which are flow counterparts of AWAC~\citep{nair2020awac}, Diffusion-QL~\citep{wang2023diffusion}, and IDQL~\citep{hansenestruch2023idql}, respectively. In the offline-to-online setting, we additionally compare with Cal-QL~\citep{nakamoto2023calql} and RLPD~\citep{ball2023efficient}. Baseline results are taken from prior work under the same benchmark protocols.

\paragraph{Evaluation protocol.} 
We follow the evaluation protocol of \citet{park2025flow} to ensure direct comparability with previously reported results. 
CoDrift is trained for a fixed number of gradient steps and evaluated periodically using 50 rollout episodes. 
Following \citet{kurenkov2022showing}, we do not select the best-performing evaluation checkpoint, which can introduce selection bias. 
For OGBench, we report the average over the final three evaluation epochs and measure task success rate. 
For D4RL, we report the final evaluation epoch, using success rate for AntMaze and normalized return for Adroit~\citep{fu2020d4rl}. 
CoDrift results are averaged over eight random seeds, except for pixel-based tasks, which use four seeds. 
For the offline-to-online experiments, we report performance after offline pretraining at $10^6$ steps and after the subsequent online fine-tuning phase at $2{\times}10^6$ total steps.

Following prior work~\citep{park2025flow,mu2026deflow}, values within 95\% of the best result in each task category are highlighted as near-best. For average rank, where lower is better, only the best result is highlighted.

\subsection{Comparison with the State of the Art}

\begin{table*}[t]
\caption{
\textbf{Offline RL results.}
CoDrift is evaluated against representative Gaussian-, diffusion-, and
flow-based offline RL methods across OGBench and D4RL.
Results for CoDrift are averaged over 8 seeds (4 for pixel-based tasks),
and baseline values are taken from prior work
\citep{tarasov2023revisiting,hansenestruch2023idql,
chen2024srpo,park2025flow,mu2026deflow}.
The pixel-based row is excluded from ranking as several baselines are unavailable.
Lower average rank is better. 
}
\label{tab:offline_rl_results}

\centering
\begingroup

\setlength{\tabcolsep}{5.5pt}
\renewcommand{\arraystretch}{1.05}

\resizebox{\textwidth}{!}{%
\begin{tabular}{@{}l*{12}{c}@{}}
\toprule

\multicolumn{1}{c}{}
&
\multicolumn{3}{c}{\texttt{Gaussian Policies}}
&
\multicolumn{3}{c}{\texttt{Diffusion Policies}}
&
\multicolumn{5}{c}{\texttt{Flow Policies}}
&
\multicolumn{1}{c}{\texttt{Drift Policy}}
\\

\cmidrule(lr){2-4}
\cmidrule(lr){5-7}
\cmidrule(lr){8-12}
\cmidrule(lr){13-13}

\texttt{Task Category}
& \texttt{BC}
& \texttt{IQL}
& \texttt{ReBRAC}
& \texttt{IDQL}
& \texttt{SRPO}
& \texttt{CAC}
& \texttt{FAWAC}
& \texttt{FBRAC}
& \texttt{IFQL}
& \fqlhead
& \deflowhead
& \ourshead
\\

\midrule

\texttt{OGBench antmaze-large}
& \scorecell{11}{1}
& \scorecell{53}{3}
& \bestscorecell{81}{5}
& \scorecell{21}{5}
& \scorecell{11}{4}
& \scorecell{33}{4}
& \scorecell{6}{1}
& \scorecell{60}{6}
& \scorecell{28}{5}
& \scorecell{79}{3}
& \bestscorecell{81}{3}
& \bestscorecell{84}{3}
\\

\texttt{OGBench antmaze-giant}
& \scorecell{0}{0}
& \scorecell{4}{1}
& \scorecell{26}{8}
& \scorecell{0}{0}
& \scorecell{0}{0}
& \scorecell{0}{0}
& \scorecell{0}{0}
& \scorecell{4}{4}
& \scorecell{3}{2}
& \scorecell{9}{6}
& \scorecell{12}{5}
& \bestscorecell{46}{22}
\\

\texttt{OGBench humanoidmaze-medium} 
& \scorecell{2}{1}
& \scorecell{33}{2}
& \scorecell{22}{8}
& \scorecell{1}{0}
& \scorecell{1}{1}
& \scorecell{53}{8}
& \scorecell{19}{1}
& \scorecell{38}{5}
& \scorecell{60}{14}
& \scorecell{58}{5}
& \scorecell{48}{4}
& \bestscorecell{83}{5}
\\

\texttt{OGBench humanoidmaze-large}
& \scorecell{1}{0}
& \scorecell{2}{1}
& \scorecell{2}{1}
& \scorecell{1}{0}
& \scorecell{0}{0}
& \scorecell{0}{0}
& \scorecell{0}{0}
& \scorecell{2}{0}
& \scorecell{11}{2}
& \scorecell{4}{2}
& \scorecell{5}{2}
& \bestscorecell{15}{5}
\\

\texttt{OGBench antsoccer-arena}
& \scorecell{1}{0}
& \scorecell{8}{2}
& \scorecell{0}{0}
& \scorecell{12}{4}
& \scorecell{1}{0}
& \scorecell{2}{4}
& \scorecell{12}{0}
& \scorecell{16}{1}
& \scorecell{33}{6}
& \scorecell{60}{2}
& \bestscorecell{67}{3}
& \bestscorecell{66}{5}
\\

\texttt{OGBench visual manipulation}
& \pendingcell
& \scorecell{42}{4}
& \scorecell{60}{2}
& \pendingcell
& \pendingcell
& \pendingcell
& \pendingcell
& \scorecell{22}{2}
& \scorecell{50}{5}
& \scorecell{65}{2}
& \bestscorecell{69}{2}
& \scorecell{61}{7}
\\

\texttt{OGBench puzzle-3x3}
& \scorecell{2}{0}
& \scorecell{9}{1}
& \scorecell{21}{1}
& \scorecell{10}{2}
& \scorecell{18}{1}
& \scorecell{19}{0}
& \scorecell{6}{2}
& \scorecell{14}{4}
& \scorecell{19}{1}
& \scorecell{30}{1}
& \scorecell{43}{4}
& \bestscorecell{82}{13}
\\

\texttt{OGBench puzzle-4x4}
& \scorecell{0}{0}
& \scorecell{7}{1}
& \scorecell{14}{1}
& \bestscorecell{29}{3}
& \scorecell{10}{3}
& \scorecell{15}{3}
& \scorecell{1}{0}
& \scorecell{13}{1}
& \scorecell{25}{5}
& \scorecell{17}{2}
& \scorecell{11}{2}
& \scorecell{22}{4}
\\

\texttt{OGBench cube-single}
& \scorecell{5}{1}
& \scorecell{83}{3}
& \scorecell{91}{2}
& \bestscorecell{95}{2}
& \scorecell{80}{5}
& \scorecell{85}{9}
& \scorecell{81}{4}
& \scorecell{79}{7}
& \scorecell{79}{2}
& \bestscorecell{96}{1}
& \bestscorecell{96}{2}
& \bestscorecell{92}{3}
\\

\texttt{OGBench cube-double}
& \scorecell{2}{1}
& \scorecell{7}{1}
& \scorecell{12}{1}
& \scorecell{15}{6}
& \scorecell{2}{1}
& \scorecell{6}{2}
& \scorecell{5}{2}
& \scorecell{15}{3}
& \scorecell{14}{3}
& \scorecell{29}{2}
& \scorecell{40}{5}
& \bestscorecell{49}{10}
\\

\texttt{OGBench scene}
& \scorecell{5}{1}
& \scorecell{28}{1}
& \scorecell{41}{3}
& \scorecell{46}{3}
& \scorecell{20}{1}
& \scorecell{40}{7}
& \scorecell{30}{3}
& \scorecell{45}{5}
& \scorecell{30}{3}
& \scorecell{56}{2}
& \scorecell{51}{3}
& \bestscorecell{59}{2}
\\

\texttt{D4RL antmaze} 
& \meanscore{17}
& \meanscore{57}
& \meanscore{78}
& \meanscore{79}
& \meanscore{74}
& \scorecell{30}{3}
& \scorecell{44}{3}
& \scorecell{64}{7}
& \scorecell{65}{7}
& \bestscorecell{84}{3}
& \bestscorecell{83}{2}
& \scorecell{73}{6}
\\

\texttt{D4RL adroit} 
& \meanscore{48}
& \meanscore{53}
& \bestmeanscore{59}
& \scorecell{52}{1}
& \scorecell{51}{1}
& \scorecell{43}{2}
& \scorecell{48}{1}
& \scorecell{50}{2}
& \scorecell{52}{1}
& \scorecell{52}{1}
& \scorecell{52}{1}
& \scorecell{50}{3}
\\

\midrule

\texttt{Average rank} $\downarrow$ (12 categories)
& \meanscore{10.96}
& \meanscore{7.38}
& \meanscore{5.29}
& \meanscore{6.21}
& \meanscore{9.42}
& \meanscore{8.04}
& \meanscore{9.79}
& \meanscore{6.58}
& \meanscore{5.58}
& \meanscore{3.08}
& \meanscore{3.13}
& \bestmeanscore{2.54}
\\

\bottomrule
\end{tabular}%
}

\endgroup
\end{table*}

\begin{table*}[t]
\caption{
\textbf{Offline-to-online RL results.}
Performance before and after online fine-tuning on the 15-task benchmark suite.
CoDrift results are averaged over 8 seeds, and baseline values are taken from prior work.
Lower average rank is better.
}
\label{tab:offline_to_online_results}

\centering
\begingroup

\setlength{\tabcolsep}{4.2pt}
\renewcommand{\arraystretch}{1.08}

\resizebox{\textwidth}{!}{%
\begin{tabular}{@{}l*{8}{c}@{}}
\toprule

\texttt{Task}
& \texttt{IQL}
& \texttt{ReBRAC}
& \texttt{Cal-QL}
& \texttt{RLPD}
& \texttt{IFQL}
& \fqlhead
& \deflowhead
& \ourshead
\\

\midrule

\texttt{humanoidmaze-medium-navigate-singletask-v0}
& \otocell{21}{13}{16}{8}
& \otocell{16}{20}{1}{1}
& \otocell{0}{0}{0}{0}
& \otocell{0}{0}{8}{10}
& \otocell{56}{35}{82}{20}
& \otocell{12}{7}{22}{12}
& \otocell{13}{5}{65}{12}
& \bestotocell{94}{5}{99}{1}
\\

\texttt{antsoccer-arena-navigate-singletask-v0}
& \otocell{2}{1}{0}{0}
& \otocell{0}{0}{0}{0}
& \otocell{0}{0}{0}{0}
& \otocell{0}{0}{0}{0}
& \otocell{26}{15}{39}{10}
& \bestotocell{28}{8}{86}{5}
& \bestotocell{44}{8}{86}{3}
& \bestotocell{52}{7}{86}{5}
\\

\texttt{cube-double-play-singletask-v0}
& \otocell{0}{1}{0}{0}
& \otocell{6}{5}{28}{28}
& \otocell{0}{0}{0}{0}
& \otocell{0}{0}{0}{0}
& \otocell{12}{9}{40}{5}
& \otocell{40}{11}{92}{3}
& \otocell{48}{13}{93}{4}
& \bestotocell{65}{10}{100}{0}
\\

\texttt{scene-play-singletask-v0}
& \otocell{14}{11}{10}{9}
& \bestotocell{55}{10}{100}{0}
& \otocell{1}{2}{50}{53}
& \bestotocell{0}{0}{100}{0}
& \otocell{0}{1}{60}{39}
& \bestotocell{82}{11}{100}{1}
& \bestotocell{58}{14}{100}{1}
& \bestotocell{92}{6}{100}{0}
\\

\texttt{puzzle-4x4-play-singletask-v0}
& \otocell{5}{2}{1}{1}
& \otocell{8}{4}{14}{35}
& \otocell{0}{0}{0}{0}
& \bestotocell{0}{0}{100}{1}
& \otocell{23}{6}{19}{33}
& \otocell{8}{3}{38}{52}
& \bestotocell{4}{4}{100}{0}
& \otocell{12}{5}{88}{32}
\\

\midrule

\texttt{antmaze-umaze-v2}
& \bestotomean{77}{96}
& \otomean{98}{75}
& \bestotomean{77}{100}
& \bestotocell{0}{0}{98}{3}
& \bestotocell{94}{5}{96}{2}
& \bestotocell{97}{2}{99}{1}
& \bestotocell{96}{3}{99}{2}
& \bestotocell{90}{5}{99}{1}
\\

\texttt{antmaze-umaze-diverse-v2}
& \otomean{60}{64}
& \bestotomean{74}{98}
& \bestotomean{32}{98}
& \otocell{0}{0}{94}{5}
& \otocell{69}{20}{93}{5}
& \bestotocell{79}{16}{100}{1}
& \bestotocell{87}{6}{99}{1}
& \bestotocell{66}{13}{98}{2}
\\

\texttt{antmaze-medium-play-v2}
& \otomean{72}{90}
& \bestotomean{88}{98}
& \bestotomean{72}{99}
& \bestotocell{0}{0}{98}{2}
& \otocell{52}{19}{93}{2}
& \bestotocell{77}{7}{97}{2}
& \bestotocell{76}{5}{97}{2}
& \bestotocell{60}{27}{98}{2}
\\

\texttt{antmaze-medium-diverse-v2}
& \otomean{64}{92}
& \bestotomean{85}{99}
& \bestotomean{62}{98}
& \bestotocell{0}{0}{97}{2}
& \otocell{44}{26}{89}{4}
& \bestotocell{55}{19}{97}{3}
& \bestotocell{61}{9}{98}{2}
& \bestotocell{59}{15}{97}{2}
\\

\texttt{antmaze-large-play-v2}
& \otomean{38}{64}
& \otomean{68}{32}
& \bestotomean{32}{97}
& \bestotocell{0}{0}{93}{5}
& \otocell{64}{14}{80}{5}
& \otocell{66}{40}{84}{30}
& \bestotocell{76}{6}{95}{4}
& \otocell{41}{17}{92}{3}
\\

\texttt{antmaze-large-diverse-v2}
& \otomean{27}{64}
& \otomean{67}{72}
& \bestotomean{44}{92}
& \bestotocell{0}{0}{94}{3}
& \otocell{69}{6}{86}{5}
& \bestotocell{75}{24}{94}{3}
& \bestotocell{76}{9}{96}{2}
& \bestotocell{7}{19}{92}{4}
\\

\midrule

\texttt{pen-cloned-v1}
& \otomean{84}{102}
& \otomean{74}{138}
& \otomean{-3}{-3}
& \otocell{3}{2}{120}{10}
& \otocell{77}{7}{107}{10}
& \bestotocell{53}{14}{149}{6}
& \bestotocell{64}{13}{142}{7}
& \bestotocell{50}{6}{143}{4}
\\

\texttt{door-cloned-v1}
& \otomean{1}{20}
& \otomean{0}{102}
& \otomean{-0}{-0}
& \otocell{0}{0}{102}{7}
& \otocell{3}{2}{50}{15}
& \otocell{0}{0}{102}{5}
& \otocell{0}{0}{99}{2}
& \bestotocell{0}{0}{108}{4}
\\

\texttt{hammer-cloned-v1}
& \otomean{1}{57}
& \otomean{7}{125}
& \otomean{0}{0}
& \otocell{0}{0}{128}{29}
& \otocell{4}{2}{60}{14}
& \otocell{0}{0}{127}{17}
& \otocell{7}{3}{106}{10}
& \bestotocell{0}{0}{139}{4}
\\

\texttt{relocate-cloned-v1}
& \otomean{0}{0}
& \otomean{1}{7}
& \otomean{-0}{-0}
& \otocell{0}{0}{2}{2}
& \otocell{-0}{0}{5}{3}
& \bestotocell{0}{1}{62}{8}
& \otocell{1}{0}{35}{8}
& \otocell{0}{0}{23}{15}
\\

\midrule

\texttt{Average rank} $\downarrow$ (15 tasks)
& \meanscore{7.10}
& \meanscore{4.90}
& \meanscore{5.47}
& \meanscore{4.30}
& \meanscore{5.63}
& \meanscore{3.07}
& \meanscore{2.83}
& \bestmeanscore{2.70}
\\

\bottomrule
\end{tabular}%
}

\endgroup
\end{table*}

\paragraph{Offline RL performance.} Table~\ref{tab:offline_rl_results} summarizes the offline RL results. Across the full benchmark suite, CoDrift achieves the best overall average rank among the compared methods, indicating strong performance across diverse task domains. 
CoDrift performs particularly strongly on OGBench tasks, while remaining competitive on D4RL. 
These results show that the proposed compositional drifting formulation provides a strong alternative to existing Gaussian-, diffusion-, and flow-based policy classes across a diverse collection of offline RL problems. 

\paragraph{Offline-to-online performance.} 
Table~\ref{tab:offline_to_online_results} reports the offline-to-online results on the 15 fine-tuning tasks. 
CoDrift can be fine-tuned online without introducing a separate online-stage objective: newly collected transitions are added to the replay buffer, and training proceeds using the same approach as in the offline phase. Across the 15 tasks, CoDrift attains the best overall average rank among the compared methods. These results show that the same compositional drifting formulation remains effective when transitioning from purely offline training to online policy improvement, without requiring an algorithmic change between the two phases.

\subsection{Effectiveness of Marginal Drifting} 
\label{sec:marginal_effectiveness} 

\begin{table}[t]
\caption{
\textbf{Effectiveness of marginal drifting.}
Success rates on four representative tasks, averaged over 8 seeds ($\pm$ standard deviation).
\texttt{w/o Marg} removes the marginal behavioral field.
$\Delta$ denotes the resulting gain from marginal drifting.
}
\label{tab:ablation}

\centering
\begingroup
\setlength{\tabcolsep}{4pt}
\renewcommand{\arraystretch}{1.15}
\small
\begin{tabular}{@{}lccc@{}}
\toprule
\texttt{Task}
& \ourshead
& \texttt{w/o Marg}
& $\Delta$
\\
\midrule
\texttt{antmaze-large}       & \bestscorecell{84}{3}  & \scorecell{77}{6} & $+7$ \\
\texttt{humanoidmaze-medium} & \bestscorecell{83}{5}  & \scorecell{77}{5} & $+6$ \\
\texttt{cube-single}    & \bestscorecell{92}{3}  & \scorecell{86}{3} & $+6$ \\
\texttt{puzzle-3x3}     & \bestscorecell{82}{13} & \scorecell{74}{4} & $+8$ \\
\midrule
\texttt{Average}             & \bestmeanscore{85.3}   & \meanscore{78.5}  & $+6.8$ \\
\bottomrule
\end{tabular}

\endgroup
\end{table}
As introduced in Sec.~\ref{sec:marginal_field}, conditional drifting estimates its attractive component from the single action observed at each state, whereas marginal drifting pools actions across states and therefore constructs its drifting field from a substantially larger positive sample. The resulting batch-level signal complements the paired state--action supervision provided by the conditional field.

To isolate the contribution of marginal drifting, we compare the full CoDrift model with a variant that removes the marginal behavioral field by setting $\lambda=1$, so that all behavioral regularization is assigned to conditional drifting. 
Table~\ref{tab:ablation} reports results on four representative tasks drawn from different OGBench domains. 
Adding marginal drifting improves performance on all four tasks, with gains ranging from 6 to 8 percentage points and an average improvement of 6.8 points. 
This result supports the role of marginal drifting as a complementary behavioral signal beyond per-state conditional drifting. 

\section{Related Work} 
\paragraph{Offline RL and behavior regularization.} 
Offline RL seeks to maximize return using a fixed dataset while avoiding extrapolation errors caused by out-of-distribution actions \citep{levine2020offline,fujimoto2019off}. Existing approaches include value regularization \citep{kumar2020conservative,an2021uncertainty}, in-sample learning \citep{kostrikov2022offline,xu2023offline}, and policy regularization. The latter retains an actor--critic formulation while constraining the learned policy toward the behavior distribution, either through explicit penalties \citep{wu2019behavior,fujimoto2021minimalist,tarasov2023revisiting} or weighted regression \citep{peng2019advantage,nair2020awac}. 
CoDrift follows the same general principle of balancing behavioral fidelity and value maximization, but represents these objectives as action-space displacement fields. 
In particular, it constrains behavior at both the conditional and marginal levels and composes these behavioral fields directly with a critic-guided value field. 

\paragraph{Generative policies.} 
Expressive generative models provide a natural policy class for representing complex and multimodal action distributions. Existing generative policies have explored several modeling paradigms, including autoregressive models~\citep{kim2024openvla,wang2026blockvla}, generative adversarial networks~\citep{vuong2022dasco}, diffusion models~\citep{wang2023diffusion,chi2023diffusion}, and flow matching~\citep{park2025flow,chang2026efficientflow,wang2026meanflowql,espinosadice2025sorl}. 
CoDrift takes a different approach based on drifting models~\citep{deng2026drifting}. Drifting models learn a stochastic generator from displacement fields acting directly on generated samples and support native one-step generation. More importantly for offline RL, their displacement-based formulation provides a natural interface for combining heterogeneous learning objectives: once different objectives are expressed as action-space fields, their effects can be composed additively before training the generator. Concurrent with our work, \citet{koo2026drifting} also applies drifting models to one-step policy learning, but uses a single drifting field toward a value-reweighted target. 
In contrast, CoDrift explicitly composes multiple objective-level fields within a unified one-step generative framework.

\section{Concluding Remarks}

We introduced CoDrift, a compositional drifting framework for generative policy learning in offline reinforcement learning. 
The central idea is to represent heterogeneous learning objectives in a common language of action-space displacement fields and compose them into a single policy field. 
For offline RL, CoDrift instantiates this principle with three complementary objective-level fields: a conditional behavioral field that preserves state-dependent behavioral structure, a marginal behavioral field that captures population-level action structure, and a critic-guided value field that promotes higher-return actions. 
The resulting joint field directly defines the regression target of a one-step stochastic actor.

Empirically, CoDrift achieves strong performance across a broad collection of OGBench and D4RL tasks in both offline and offline-to-online settings, attaining the best overall average rank among the compared methods. These results demonstrate that compositional drifting provides a competitive alternative to existing Gaussian-, diffusion-, and flow-based policy classes, while retaining native one-step stochastic generation at deployment.

More broadly, we view the main advantage of the drifting formulation as its objective compositionality. Different requirements need not share the same loss construction; it is sufficient that each can be translated into a displacement in action space. Their interaction can then be expressed through the additive composition of the corresponding fields and absorbed into a single generative policy. 
We hope this perspective provides a useful foundation for incorporating additional objectives and constraints into future one-step generative policies for reinforcement learning.
\bibliographystyle{iclr2025_conference}
\bibliography{main}

\end{document}